\documentclass{article}
\usepackage{amssymb}
\usepackage{amsmath}
\usepackage{siunitx}
\usepackage{graphicx}
\usepackage{wrapfig}
\usepackage{subcaption}
\usepackage{booktabs}
\usepackage{enumitem}
\usepackage{xcolor}

\usepackage[preprint]{corl_2026}

\newcommand{\eref}[1]{Eq.~\eqref{#1}}

\usepackage{xcolor}
\newcounter{todos}
\AtEndDocument{\ifnum\value{todos}>0 \PackageWarning{TODOS}{There are \arabic{todos} todos left in this paper! Fix them before submitting the paper!} \fi}

\usepackage{bm}

\title{ConCent: Contact-Centric Real-to-Sim-to-Real Learning from One Demonstration}

\author{
  Heecheol Kim$^{1}$ \quad Namiko Saito$^{1}$ \quad Katsushi Ikeuchi$^{2}$ \quad Yasuyuki Matsushita$^{1}$ \\[6pt]
  $^{1}$Microsoft Research Asia - Tokyo \quad $^{2}$The University of Tokyo
}

\begin{document}
\maketitle

\begin{center}
  \includegraphics[width=\textwidth]{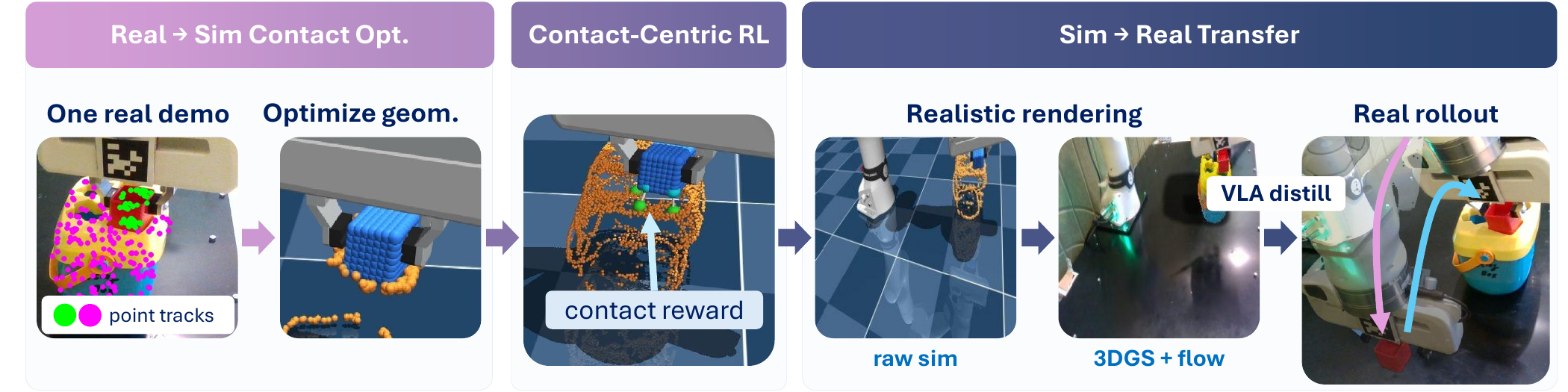}
  \captionof{figure}{\textbf{ConCent} grounds simulation in real interaction physics from a single demonstration and learns a contact-rich policy that transfers back to the real world. In \emph{Real\,$\rightarrow$\,Sim Contact Optimization}, one real RGB-D demo provides point tracks that we use to optimize \emph{contact geometry}, so the simulator reproduces the observed motion; replaying it extracts a \emph{contact event sequence} used as a structured reward. A \emph{contact-centric RL} policy is then trained with this reward. Finally, in \emph{Sim\,$\rightarrow$\,Real Transfer}, realistic rendering (raw sim and 3DGS+flow) distills the policy into a VLA for real-world deployment with no additional real data.}
  \label{fig:overview}
\end{center}

\begin{abstract}
Sim-to-real policy transfer---deploying policies trained in simulation in the real world---is a promising paradigm for scaling robot manipulation without large-scale real-world data. However, transferring simulation-trained policies remains challenging due to discrepancies in contact dynamics---particularly in contact-rich tasks where subtle differences can alter task outcomes entirely.
Because interaction between the manipulated object and the environment is mediated through contact, task success depends on accurately reproducing task-relevant contacts. Accordingly, in manipulation, contact-centric fidelity---reproducing both the contact event sequence (when, where, and how contacts occur) and the local contact dynamics (how forces and motions evolve at each contact)---is a necessary condition for task success. 
Based on this insight, we propose a contact-centric real-to-sim-to-real RL framework that uses task-relevant contact event sequences extracted from real demonstrations as the learning objective. We approximate objects as groups of primitives and optimize their contact geometry in simulation so that the resulting local contact dynamics explain the observed state transitions. The contact event sequence is automatically extracted by replaying the demonstration. This sequence serves as a structured reward signal, guiding the policy toward physically plausible contact regimes validated in reality and preventing exploitation of unrealistic simulator contacts. The signal is obtained automatically, requiring no per-task reward design.
Experiments on contact-rich manipulation tasks demonstrate more stable and robust sim-to-real policy transfer compared to unconstrained RL baselines.
\end{abstract}


\section{Introduction}
\label{sec:introduction}
Recent advances in robot manipulation have pursued scaling policy learning through large-scale real-world data collection~\citep{brohan2023rt2, kim2024openvla, black2024pi_0}. However, these approaches face a fundamental limitation: they require massive amounts of high-quality real-world data, and these data must be collected repeatedly whenever the task, scene, or robot configuration changes---an expensive and time-consuming process. Consequently, sim-to-real policy transfer, which trains policies in simulation 
and deploys them in the real world, has emerged as a compelling paradigm for scaling manipulation policies without additional real-world data collection.

In practice, however, reliably transferring simulation-trained policies to the real world remains challenging. This difficulty is particularly pronounced in contact-rich manipulation, where object interactions are primarily mediated by contact. Even subtle discrepancies in contact dynamics between simulation and reality can significantly alter object behavior and task outcomes. As a result, simulation-based RL often exploits unrealistic contact modes during training, producing policies that fail to transfer.

Prior work has addressed the sim-to-real gap through physical parameter identification~\citep{yu2017preparing, lin2025simtoreal} and domain randomization~\citep{tobin2017domain, peng2018sim}, but these approaches often fail for contact-rich manipulation. This is because the success or failure of manipulation depends not on global physical parameters, but on the local contact behavior that arises during task execution: both the spatial-temporal pattern of contact events among the robot, objects, and environment, and the local dynamics each contact induces on the objects.

\begin{wrapfigure}{r}{0.5\textwidth}
  \centering
  \vspace{-10pt}
  \includegraphics[width=0.48\textwidth]{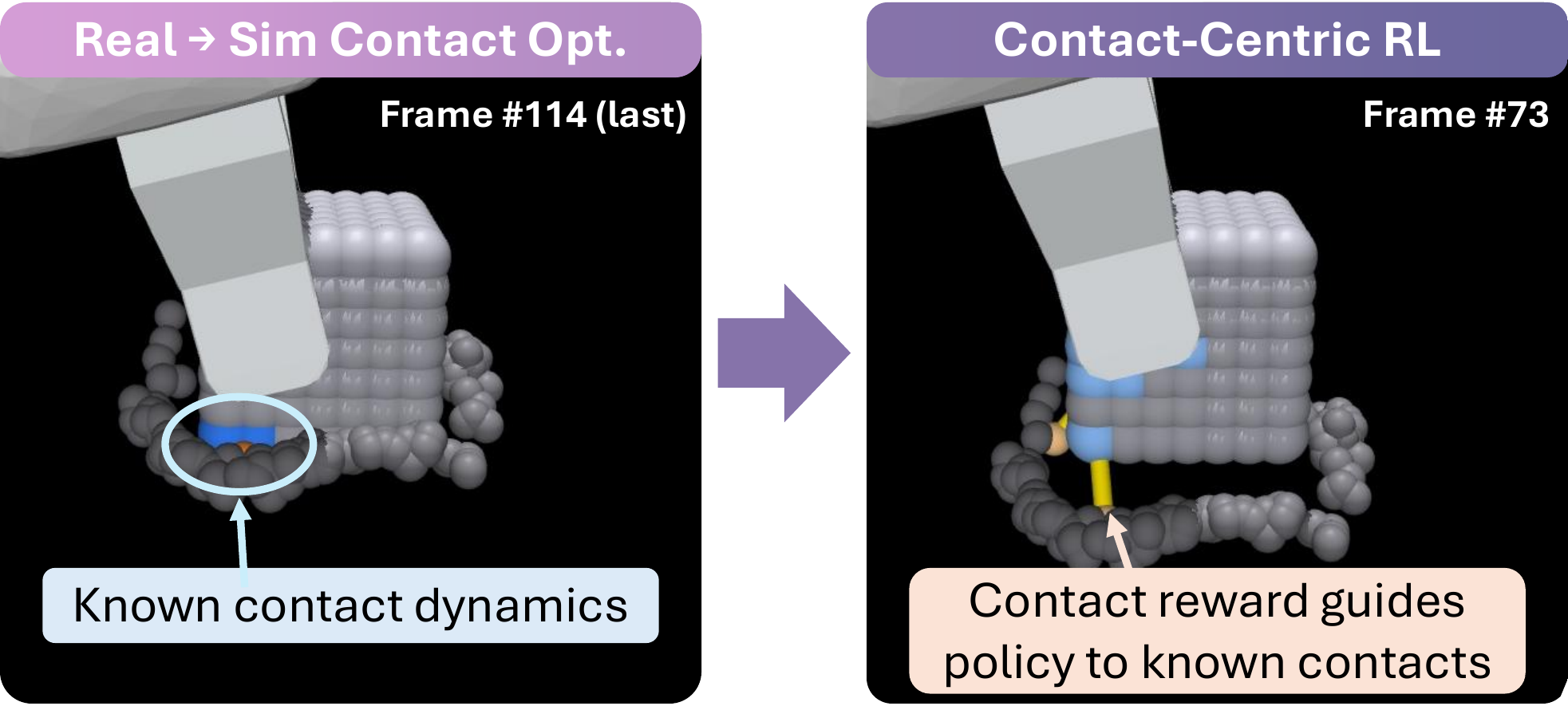}
  \caption{\textbf{Principle of ConCent.}
  The local contact dynamics can be revealed by a real-world demonstration
  (left); the contact-centric RL then constrains the policy to it, minimizing
  the sim-to-real gap (right).}
  \label{fig:principle}
  \vspace{-8pt}
\end{wrapfigure}
From this perspective, we propose a new real-to-sim-to-real RL framework (grounding the simulator in a real demonstration, training an RL policy in that simulator, and deploying it back to the real world) that focuses on task-relevant contact-centric fidelity rather than global physics matching (Fig.~\ref{fig:overview}). The key insight is that what matters in manipulation is not the faithful reproduction of all physical phenomena, but reproducing both the \emph{local contact dynamics} (how forces and motions evolve at each contact) and the \emph{contact event sequence} (when, where, and how contacts occur), which together form a necessary condition for task success. We first match the local contact dynamics by optimizing object contact geometry so that the simulator reproduces observed state transitions from a real demonstration; we then extract the resulting contact event sequence and interpret it as sequential physical constraints that restrict policy learning. This guides the policy to explore physically plausible interaction regimes consistent with real demonstrations, rather than exploiting unrealistic simulator contacts. Such constraint-based learning enables the policy to acquire manipulation strategies that remain valid in the real world, even in the presence of global physics errors in simulation (Fig.~\ref{fig:principle}).

To realize this framework, we approximate object geometry as a set of primitives and optimize their configuration in simulation so that the resulting local contact dynamics explain the observed state transitions from a single real-world demonstration. By replaying the demonstration in the contact-geometry-optimized simulation, we automatically extract the contact event sequence and use it as a reward signal for RL.
Unlike imitating the demonstration trajectory directly, this approach converts observed contact events into physical constraints that bound the policy search space, enabling the learning of closed-loop contact strategies that are robust to diverse initial conditions and perturbations. Because the reward signal is derived automatically from the demonstrated contact structure rather than hand-crafted per task, the framework generalizes across diverse manipulation tasks without manual reward engineering.

Our contact-centric formulation also yields a natural computational advantage. Because the demonstrated contact sequence explicitly identifies which primitives will participate in task-relevant contacts, only those primitives require full collision fidelity in simulation. We exploit this through \emph{Virtual Collision Penalty (VCP)}: primitives in the contact sequence are instantiated as physics bodies with full collision processing, while the remaining primitives are enforced through GPU-parallelizable distance penalties. This makes contact-centric RL simulation speed practical---an optimization unavailable to contact-agnostic RL methods, which cannot know a priori which contacts matter.

We evaluate the proposed method on a contact-rich manipulation task involving pick-and-insert operations, which exhibit multiple distinct contact phases such as grasping, alignment, and insertion, and demonstrate more stable and robust sim-to-real policy transfer than unconstrained RL baselines.

Our contributions are as follows:
\begin{enumerate}[itemsep=2pt,parsep=0pt,topsep=2pt]
    \item A real-to-sim-to-real RL framework that achieves contact-rich manipulation by optimizing local contact dynamics from a real demonstration and then learning an RL policy constrained by those contact dynamics in simulation, thereby acquiring in simulation a policy that conforms to real contact.
    \item Virtual Collision Penalty (VCP), a computational optimization naturally enabled by our contact-centric formulation: since the contact sequence identifies task-relevant primitives, only those require full collision fidelity, making scalable GPU-parallel training tractable.
    \item Empirical demonstration of robust sim-to-real policy transfer on contact-rich tasks beyond pick-and-place, including insertion, where unconstrained RL baselines fail to transfer reliably.
\end{enumerate}

\section{Related Work}
\label{sec:related_work}
\subsection{Contact Extraction from Demonstrations}

Prior work interprets demonstrations as contact modes or constraint sequences~\citep{takamatsu2007recognizing, ikeuchi2025learning, li2023augmentation, subramani2018inferring}, or leverages differentiable physics to optimize contact points from video~\citep{zhu2023difflfd}.
These works share a philosophical similarity with ours, but their objectives lie in compliant execution, planning, or model-based action generation; they do not use inferred contact structure as a learning objective for RL.
ContactGaussian-WM~\citep{wang2026contactgaussian} also optimizes Gaussian-based collision geometry from video via differentiable physics, but targets learned forward prediction within a custom differentiable engine. Our method instead optimizes contact geometry to be compatible with general-purpose, GPU-accelerated non-differentiable simulators (e.g., MuJoCo), and integrates the resulting contact structure directly into RL reward and policy learning rather than using it for state prediction alone.
Our work extends this direction by converting the contact event sequence from real demonstrations into reward signals for simulation-based policy learning.

\subsection{Demonstration-Guided RL and Reward Design}

Demonstrations have been used to scale imitation learning via data augmentation~\citep{mandlekar2023mimicgen} or as reference-motion tracking rewards for RL~\citep{peng2018deepmimic}.
Alternatively, inverse RL~\citep{ng2000irl, ho2016gail} and LLM-based reward generation~\citep{ma2024eureka} estimate or synthesize rewards from demonstrations but suffer from reward ambiguity or lack of physical grounding.
More recently, X-Sim~\citep{dan2025xsim} and Human2Sim2Robot~\citep{lum2025human2sim2robot} extract dense object pose trajectories from human demonstrations and use them as continuous tracking rewards for simulation RL. While effective for tasks where object pose progress correlates with success, these pose-tracking rewards are fundamentally contact-agnostic: they reward \emph{where} the object ends up regardless of \emph{how} contact is made or broken. For contact-rich manipulation where the same object motion can result from physically distinct (and differently transferable) contact modes, such signals provide insufficient constraint.
Rather than tracking object motion continuously, we convert physically valid contact events into sequential constraints that bound the policy search, grounding the learning objective in the physical essence of manipulation rather than kinematic progress.

\subsection{Sim-to-Real Transfer}

Physical parameter estimation~\citep{yu2017preparing} and domain randomization~\citep{tobin2017domain, peng2018sim} are widely adopted for sim-to-real transfer but have fundamental limitations when contact structures are inaccurate.
Recent real-to-sim approaches improve visual, geometric, and even physics-ready scene reconstruction~\citep{mildenhall2021nerf, kerbl20233d, moran2025splatting}, but they generally optimize scene-level fidelity rather than explicitly identifying which contacts are task-relevant for policy learning.
Real-to-sim-to-real systems such as RialTo~\citep{torne2024reconciling}, Re3Sim~\citep{han2025re3sim}, and Digital Cousins~\citep{dai2024digital} construct simulation environments from real scenes (via exact digital twins, photorealistic rendering, or semantically matched ``cousin'' assets) and train policies through RL or imitation learning for zero-shot transfer. However, these approaches target scene-level visual or geometric fidelity and do not extract or exploit task-relevant contact structure, limiting their applicability to contact-rich manipulation where subtle contact geometry determines success or failure.
For contact-rich assembly, IndustReal~\citep{tang2023industreal} achieves robust sim-to-real transfer using hand-crafted signed distance function (SDF)-based
alignment rewards and penetration-aware policy updates, given known CAD models and predefined assembly targets. In contrast, our method does not assume known target geometry or simulator-derived rewards; instead, it infers task-relevant contact anchors from real demonstrations and uses them to constrain simulation construction, reward design, and policy learning, enabling contact-rich manipulation beyond predefined assembly benchmarks.

\section{Method}
\label{sec:method}
We build a real-to-sim-to-real pipeline that takes a single real demonstration and turns it into a deployable policy (Fig.~\ref{fig:overview}): physically grounding the simulator via demonstration-extracted contact events (Sec.~\ref{sec:contact_physics}), and training a contact-centric RL policy that is distilled into a vision-language-action model for real-world deployment (Sec.~\ref{sec:policy_learning}).

\subsection{Bridging the Contact Physics Gap via Demonstration-Extracted Contact Events}
\label{sec:contact_physics}

Our framework grounds the simulator in real interaction physics by extracting, from a real demonstration, a structured description of \emph{what contacts actually occur} and using it as the reference that subsequent simulation and policy learning must respect. We achieve this in two steps. First, we optimize an explicit contact geometry so that the simulator reproduces the demonstrated object motion (Sec.~\ref{sec:contact_primitive}). Second, we replay the demonstration through this contact geometry to read off a deterministic contact event sequence (Sec.~\ref{sec:contact_extraction}). The resulting sequence anchors the rest of the method: it defines the physically validated interaction regime that the policy is later trained to reproduce, and prevents RL from exploiting contacts that never occur in the real environment.

\subsubsection{Contact Geometry Optimization}
\label{sec:contact_primitive}

Defining a contact event sequence requires an explicit contact representation in simulation that can explain the physical interactions observed in the real demonstration. We refer to this representation as the \emph{contact geometry}, defined as a set of contact primitives $C = \{c_i\}_{i=1}^N$ that serve as actual collision geometries within the simulator. Each primitive $c_i$ possesses local physical properties including position in $\mathbb{R}^3$, mass, and friction coefficient, and directly participates in the simulator's contact solver.

From the real demonstration, we apply SAM2~\citep{ravi2024sam2} and TAPIR~\citep{doersch2023tapir} to RGB-D observations to extract the observed object point cloud track $V = \{v_j\}_{j=1}^M$.

The optimization objective for the contact geometry is to find $C$ such that the
simulator best explains the observed object point cloud track. We replay the demonstrated actions in simulation, so the object point cloud track $V_t$ at time $t$ predicted by the simulator is a function $f(a_{1:t}; C)$ of the replayed actions and the contact geometry:

\begin{equation} \label{eq:contact_optimization}
C^* = \arg\min_{C} \sum_{t} D_{\mathrm{CD}}\!\big( V_t,\; f(a_{1:t}; C) \big).
\end{equation}
Here \hbox{$a_{1:t}=(a_1,\dots,a_t)$} is the replayed action sequence up to $t$, and $D_{\mathrm{CD}}(\cdot,\cdot)$ is the Chamfer distance.
We optimize $C$ via a sampling-based evolutionary strategy, iteratively replaying the demonstration in simulation and updating $C$ to minimize the error in \eref{eq:contact_optimization} (implementation details in Appendix~\ref{sec:appendix_es}). The resulting $C^*$ constitutes the contact geometry used throughout our framework; all subsequent simulation-based analysis and policy learning are built upon this representation.

\subsubsection{Contact Event Sequence Extraction}
\label{sec:contact_extraction}

Given the optimized contact geometry $C^*$, we replay the real demonstration in simulation to track contacts between robot links and contact primitives, as well as among contact primitives themselves, at each timestep. For each contact, we record the occurrence time and the involved entities (robot link or primitive ID).

The resulting sequence is defined as the contact event sequence, where $S$ is the number of stages given by the demonstration's contact-phase transitions (Appendix~\ref{sec:appendix_staging}):
\begin{equation}
E = (e_1, e_2, \dots, e_S).
\end{equation}
Each event $e_s$ contains the set of robot link and primitive IDs in contact at that timestep. These are explicitly detectable quantities in physics simulators such as MuJoCo~\citep{todorov2012mujoco}, providing a deterministic and structured representation of the contact structure from the demonstration.

The extracted $E$ defines a physically plausible interaction regime validated in the real environment and serves as the sequential objective for policy learning in the following section.

\subsection{Sim-to-Real Policy Transfer via Contact-Centric RL}
\label{sec:policy_learning}

Given the extracted contact event sequence, we train an RL policy that uses this sequence as a structured reward signal (Sec.~\ref{sec:ppo}), accelerate training by pruning non-essential collisions via a virtual collision penalty (Sec.~\ref{sec:vcp}), and distill the resulting policy into a vision-language-action model for real-world deployment (Sec.~\ref{sec:vla}).

\subsubsection{Contact-Centric Policy Optimization}
\label{sec:ppo}

We train the policy $\pi_\theta$ using PPO~\citep{schulman2017ppo} implemented in Brax~\citep{freeman2021brax}. The observation includes joint encodings, end-effector pose, object positions, and per-contact-pair features (contact point, displacement, approach direction, target normal). The action consists of 7-DoF delta-joint commands and a gripper command. The reward combines a reaching term $r_t^{\text{reach}}$ that drives the active contact pairs together, action/velocity and object-disturbance regularization $r_t^{\text{reg}}$, and the VCP penalty $r_t^{\text{vcp}}$:
\begin{equation}
    r_t = \lambda_{\text{reach}}\, r_t^{\text{reach}} + \lambda_{\text{reg}}\, r_t^{\text{reg}} + \lambda_{\text{vcp}}\, r_t^{\text{vcp}},
    \label{eq:reward}
\end{equation}
where $\lambda_\bullet$ weight the three terms, defined as follows.
The \emph{reaching} term \hbox{$r_t^{\text{reach}}=\exp(-\max_i d_t^{(i)}/\sigma)$} drives the active contact pairs together, where $d_t^{(i)}$ is the distance of the $i$-th active contact pair; using the worst (maximum-distance) pair rather than the mean forces the policy to close all demonstrated contacts evenly instead of collapsing a subset.
The \emph{regularization} term $r_t^{\text{reg}}$ is an $L_2$ penalty on the action and joint velocity, plus a differential penalty against unnecessarily pushing or rotating the manipulated objects.
The VCP term $r_t^{\text{vcp}}$ (Sec.~\ref{sec:vcp}) is a distance-based penalty on non-essential robot--object and object--object primitive pairs, with early termination on severe penetration.
To facilitate exploration, we diversify initial states by sampling poses from the demonstration while preserving the current contact chain, inspired by OmniReset~\citep{yin2026omnireset} (details in Appendix~\ref{sec:appendix_reset}). Full observation/action specifications, per-term weights $\lambda_*$, and hyperparameters are provided in Appendix~\ref{sec:appendix_ppo}.

\subsubsection{Virtual Collision Penalty: Pruning Collisions for Scalable RL}
\label{sec:vcp}

Representing each object as a group of sphere primitives faithfully captures fine-grained contact dynamics, but inflates the number of collidable bodies the simulator must process by orders of magnitude, and the per-step cost of collision detection and contact force resolution quickly dominates training time. Reducing this cost without sacrificing the fidelity of \emph{task-relevant} contacts is therefore essential for practical RL. The contact event sequence extracted in Sec.~\ref{sec:contact_extraction} provides exactly the information needed: it tells us which primitives ever participate in a real contact, and which do not.

Exploiting this, 
VCP instantiates only the primitives that appear in the demonstrated contact sequence as actual simulation bodies (a drop-in modification to existing simulators such as MuJoCo~\citep{todorov2012mujoco}), confining expensive collision detection and contact force computation to this small, task-relevant subset. The remaining primitives, never observed in the real contact event sequence, are omitted from the physics solver entirely, drastically reducing the number of bodies the simulator must process.

During training, non-collidable primitives serve a collision avoidance role via reward penalties rather than precise collision responses. We compute the distance between non-collidable primitives and target meshes, detecting a collision when this distance falls below the sphere radius, and impose a penalty on the policy reward. Although such penalties never inject contact forces into the simulation, the policy is incentivized to learn collision avoidance behavior at convergence. This distance-based approach is highly parallelizable on GPUs and eliminates expensive contact force computation.

\subsubsection{Sim-to-Real Deployment via VLA Distillation}
\label{sec:vla}

The trained RL policy operates on object representations within simulation and thus cannot be directly deployed in the real world. To bridge the appearance gap, we combine 3D Gaussian Splatting (3DGS)~\cite{kerbl20233d} for objects (which require explicit representations compatible with the physics simulator) with a U-Net-based flow-matching generative model that renders the robot and background one frame at a time, conditioned on the current simulation render, which captures non-parameterizable appearance factors (e.g., shadows, cable deformations). Architecture details are described in Appendix~\ref{sec:appendix_visual}.

For real-world deployment, we train a Vision-Language-Action (VLA) model using synthetic data generated by the RL policy. Specifically, we store simulation rollouts from the RL policy, apply the hybrid rendering pipeline to add photorealistic object and robot/background rendering, and use this data to train the VLA. We adopt FLOWER~\citep{li2025flower}, a lightweight and efficient VLA architecture.

\section{Experiments}
\label{sec:result}

\subsection{Experimental Setup}
\label{subsec:exp_setup}

We evaluate ConCent on a \emph{shape sorter} insertion task.
The task consists of three sequential stages (\emph{picking} the block,
\emph{aligning} it over the hole, and \emph{inserting} it), where a
$\qty{40}{\milli\meter}$ block must be fit into a $\qty{42}{\milli\meter}$ square hole,
leaving a very tight clearance of about $\qty{2}{\milli\meter}$ per side.
In particular, if the block is misaligned during insertion, contact with the
hole's edge causes the block to lose its pose irrecoverably and the entire task
fails; success therefore depends heavily on whether the simulation reproduces
the same \emph{local contact} as the real world, making this a representative
contact-rich task.
The initial contact geometry of each object is obtained by uniformly sampling
the Gaussian means of a learned 3D Gaussian Splatting model via Farthest Point
Sampling to approximate the overall shape, and instantiating these as sphere
geometries within the simulation (Appendix~\ref{sec:appendix_visual}).

The experiments in this section are designed to answer the following three questions.
\begin{enumerate}[itemsep=2pt,parsep=0pt,topsep=2pt]
    \item[\textbf{Q1.}] Does ConCent learn a reliable manipulation policy on a precision insertion task,
    and are the individual contact-centric components (contact geometry
    optimization, contact-event reward) essential for this?
    (Sec.~\ref{subsec:main_results})
    \item[\textbf{Q2.}] How do the two contact-centric components produce
    the gains observed in Q1? (Sec.~\ref{subsec:fidelity_results})
    \item[\textbf{Q3.}] Does VCP reduce the computational cost of training and
    scale to massively parallel simulation, compared to full physical contact?
    (Sec.~\ref{subsec:vcp_results})
\end{enumerate}

\subsection{Q1: Real-World Manipulation Performance of ConCent}
\label{subsec:main_results}

To verify whether our method transfers to real-world
precision insertion, and whether each component contributes to this, we compare it
against three ablations that remove its core components.
\begin{itemize}[itemsep=2pt,parsep=0pt,topsep=2pt]
    \item \textbf{w/o Contact Geometry Optimization:} removes the step that optimizes
    the contact geometry from the demonstration to align local contact dynamics
    (Sec.~\ref{sec:contact_primitive}), using the initial contact geometry as is.
    \item \textbf{w/o Contact-Event Reward:} replaces the proposed contact-event-based
    reward with a hand-designed, geometry-based insertion reward
    (align-$xy$ $+$ depth-$z$ $+$ success bonus)
    (Appendix~\ref{sec:appendix_engineered_reward}).
    \item \textbf{Unconstrained RL:} removes both components above (no contact geometry
    optimization $+$ hand-designed reward), corresponding to an RL baseline with no
    contact-centric constraint at all.
\end{itemize}
Since the difference between conditions is most pronounced in the \emph{insertion}
stage, which requires precision contact, all conditions share the same
picking and alignment policy generated by our method, and we ablate only the
insertion stage. 
For a fair comparison, we apply the same 20 distinct initial poses (block and box
poses) extracted from the dataset to all conditions, measure the insertion success of
each rollout, and report them in Table~\ref{tab:main_results}.

\begin{wraptable}{r}{0.5\textwidth}
    \centering
    \vspace{-10pt}
    \caption{Ablation on the success rate of the shape
    sorter task ($N=20$). All conditions share the same proposed picking and alignment ($100.0\%$) stage; only the insertion stage is ablated. Values are insertion success rates (\%), with (success/total) in parentheses.}
    \label{tab:main_results}
    \setlength{\tabcolsep}{4pt}
    \renewcommand{\arraystretch}{1.2}
    \footnotesize
    \begin{tabular}{l c}
        \toprule
        \textbf{Method} & \textbf{Succ. (\%)} \\
        \midrule
        \textbf{ConCent} & \shortstack{\textbf{80.0} \scriptsize(16/20)} \\
        \quad w/o Contact Geom. Optim. & \shortstack{20.0 \scriptsize(4/20)} \\
        \quad w/o Contact-Event Reward & \shortstack{50.0 \scriptsize(10/20)} \\
        \quad Unconstrained RL & \shortstack{30.0 \scriptsize(6/20)} \\
        \bottomrule
    \end{tabular}
    \vspace{-8pt}
\end{wraptable}

\begin{figure*}[t]
    \centering
    \includegraphics[width=\textwidth]{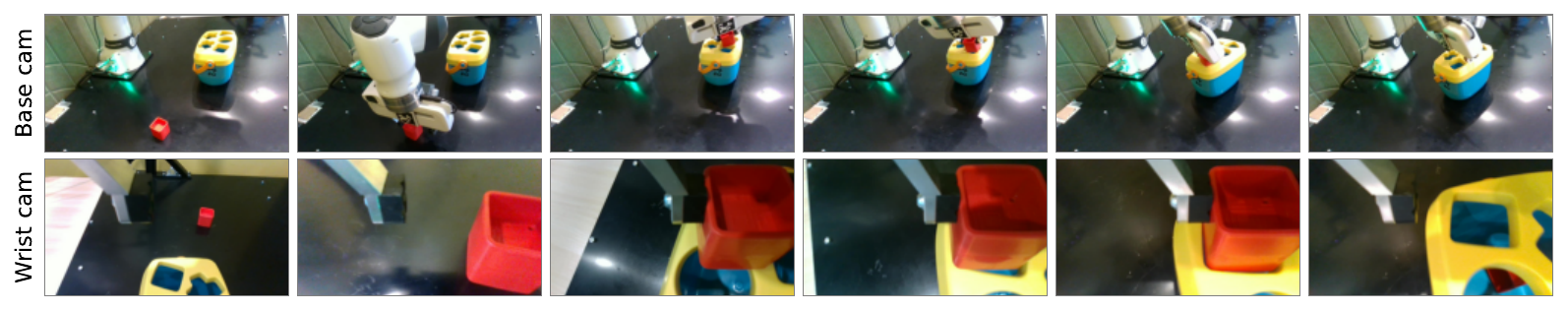}
    \caption{\textbf{Real-world rollout keyframes (ConCent).}
    A successful rollout.
    The block is grasped, aligned, and inserted into the tight
    $\qty{2}{\milli\meter}$-clearance hole; in the last frame
    the block has vanished into the hole.}
    \label{fig:rollout_filmstrip}
\end{figure*}

Our method achieves an end-to-end success rate of \textbf{80.0\%} on
precision insertion (Table~\ref{tab:main_results}), showing that the policy trained in simulation transfers reliably
to the real world (Fig.~\ref{fig:rollout_filmstrip}).
The two contact-centric components work best in tandem, substantially above any ablation (20.0--50.0\%). In particular,
without contact geometry optimization the contact-event reward shows no clear
advantage over a hand-designed reward (20.0\% vs.\ 30.0\%/50.0\%), suggesting
that the contact-event reward is effective mainly when paired with optimized
contact geometry that supplies accurate contact anchors.
This shows that both contact-centric components are essential for the sim-to-real policy transfer of precision insertion.

\begin{figure}[tb]
    \centering
    \includegraphics[width=0.48\textwidth]{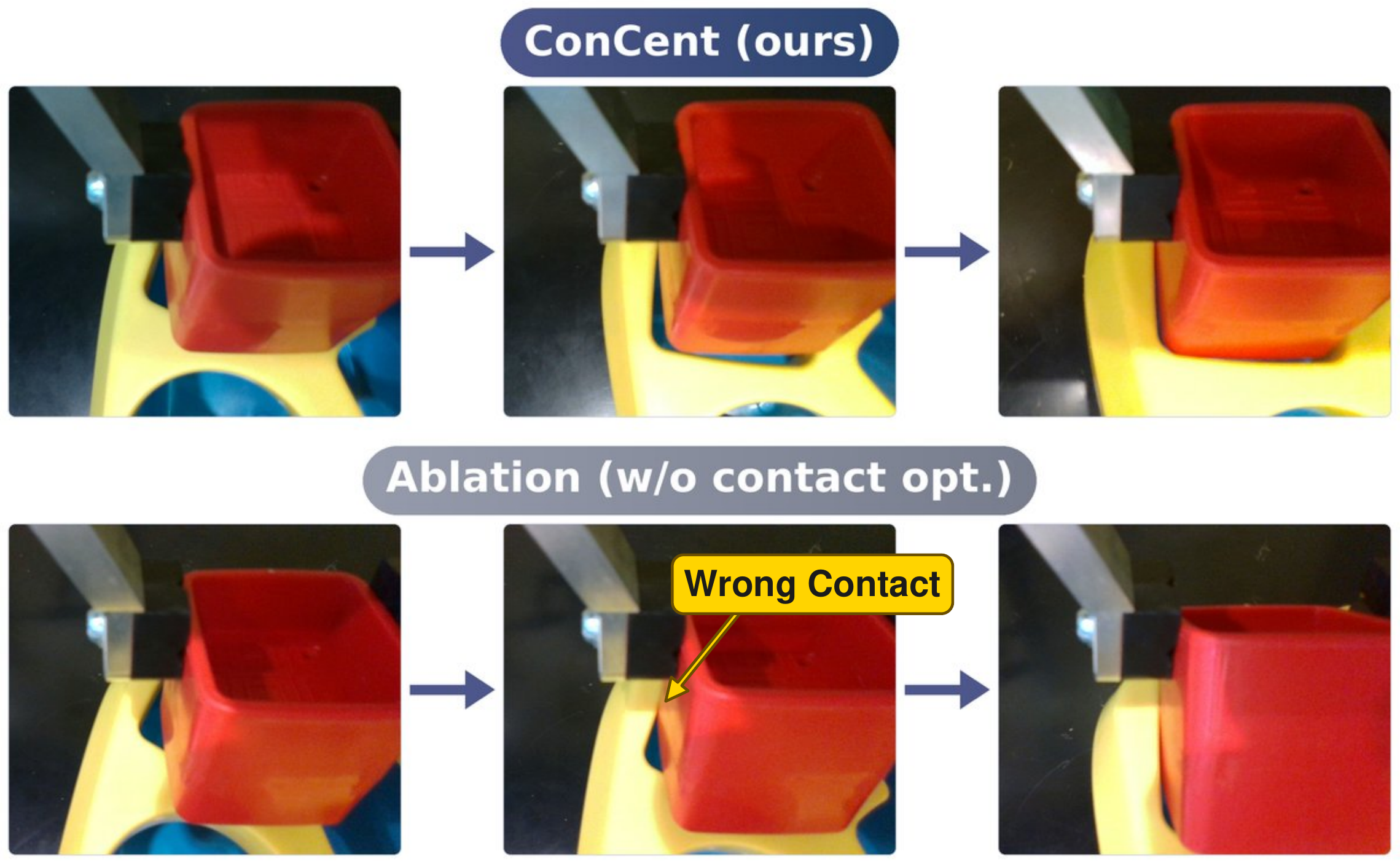}
    \caption{\textbf{In-hand view during insertion} (wrist camera).
    \textbf{Top (ConCent):} the block stays aligned and slides in.
    \textbf{Bottom (w/o contact optimization):} a \emph{Wrong Contact} knocks
    the in-hand pose off and the block jams against the rim.}
    \label{fig:inhand_compare}
\end{figure}
In particular, the effect of contact geometry optimization is observable in
the Fig.~\ref{fig:inhand_compare}. Thanks to the
optimized contact geometry, ConCent keeps the block aligned with the hole and inserts
it smoothly without colliding with the rim, whereas the ablation without contact
geometry optimization causes the block to collide with and jam against the hole edge,
lose its pose, and fail the insertion.
That is, success on this task is governed by the
reproduction of task-relevant local contact (Q1).

\subsection{Q2: How the Two Contact-Centric Components Work}
\label{subsec:fidelity_results}

To understand the gains in Q1, we analyze how each of the two
contact-centric components mechanistically contributes: contact geometry
optimization, which aligns the simulator's local contact dynamics with the
demonstration, and the contact-event reward, which imposes the demonstrated
contact schedule as a structured learning objective.

\begin{figure}[t]
    \centering
    \begin{subfigure}[b]{0.34\textwidth}
        \centering
        \includegraphics[width=\textwidth]{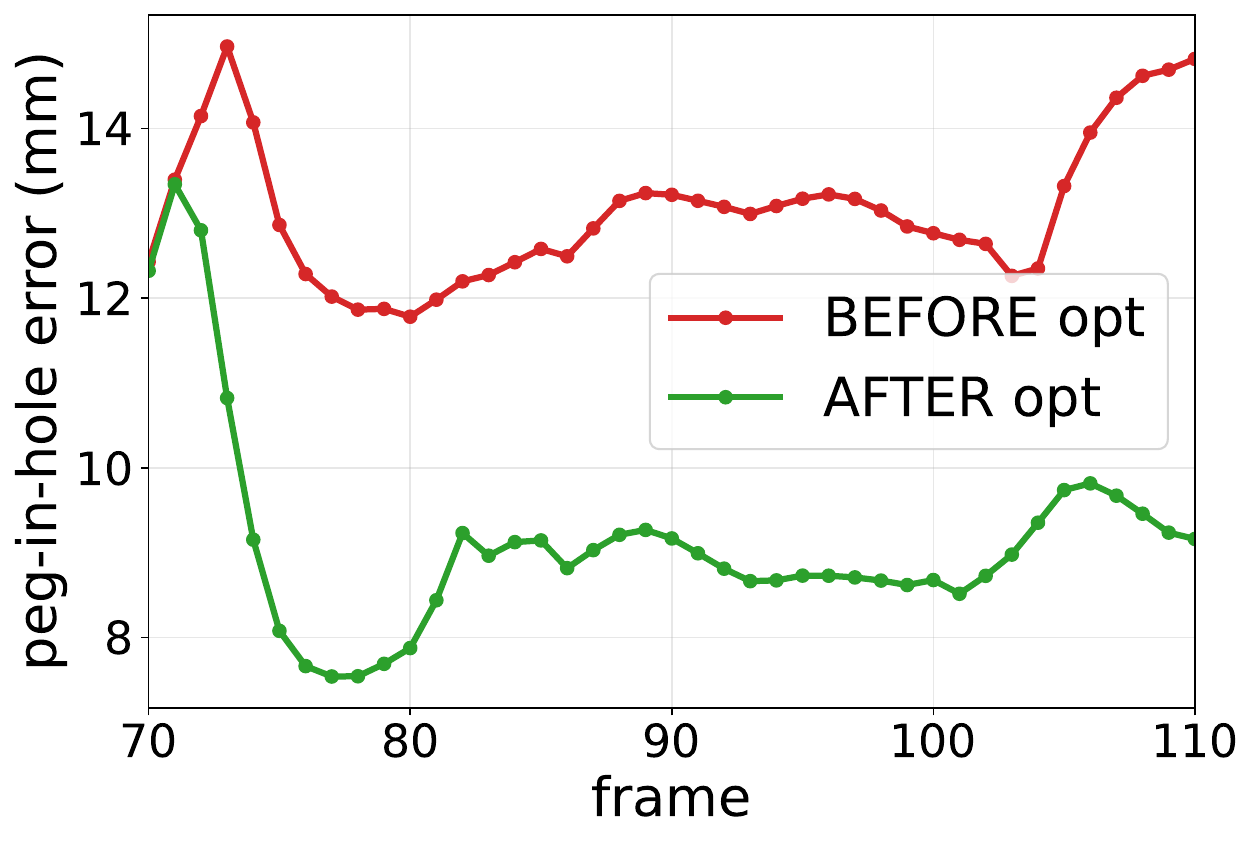}
        \caption{Block position error}
        \label{fig:contact_opt_err}
    \end{subfigure}
    \hspace{0.02\textwidth}
    \begin{subfigure}[b]{0.39\textwidth}
        \centering
        \includegraphics[width=\textwidth]{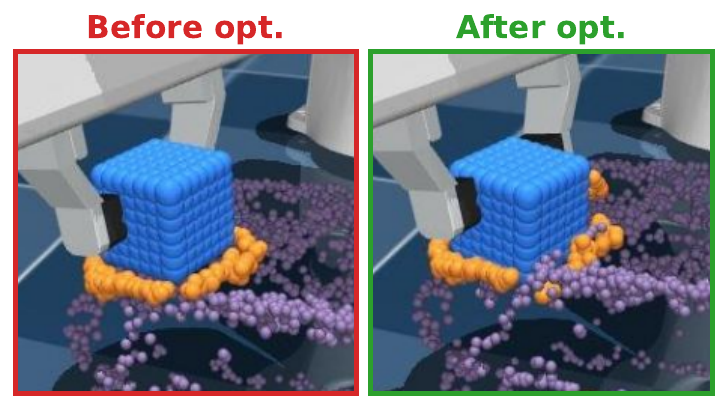}
        \caption{Replayed simulation frames}
        \label{fig:contact_opt_frames}
    \end{subfigure}
    \caption{\textbf{Contact geometry optimization aligns the simulated insertion
    with the demonstration.}
    \emph{(a)} Block position error (simulation vs.\ real observation) over the
    insertion phase, before vs.\ after optimization.
    \emph{(b)} The last replayed frame under the initial contact geometry
    (left, red) and the optimized geometry $C^*$ (right, green).
    }
    \label{fig:contact_opt}
\end{figure}

\paragraph{Contact geometry optimization aligns local contact dynamics.}
To understand why removing contact geometry optimization in Q1 substantially
degraded insertion performance, we measure how well the optimization actually
aligns the simulation's contact dynamics with the demonstration.
We replay the same recorded robot actions through the contact geometry
before and after optimization (\eref{eq:contact_optimization})
and track the block's position error between simulation and the real
observation throughout the insertion phase
(Fig.~\ref{fig:contact_opt_err}).
Before optimization, the initial contact geometry fails to explain the real
contacts: the block slips against the rim instead of seating, leaving
a large block position error (mean $13.1$\,mm). After
optimization the simulation reproduces the demonstrated insertion, dropping the
error to $9.1$\,mm on average.
Figure~\ref{fig:contact_opt_frames} shows the last replay frame: with the initial
geometry the block fails to seat into the hole (left), whereas the optimized
geometry $C^*$ drives it into the hole exactly as demonstrated (right). Notably,
what the optimization aligns is the
\emph{task-relevant relative contact} between block and hole, which is precisely
what governs real-world insertion success.

\begin{figure}[t]
    \centering
    \includegraphics[width=0.75\textwidth]{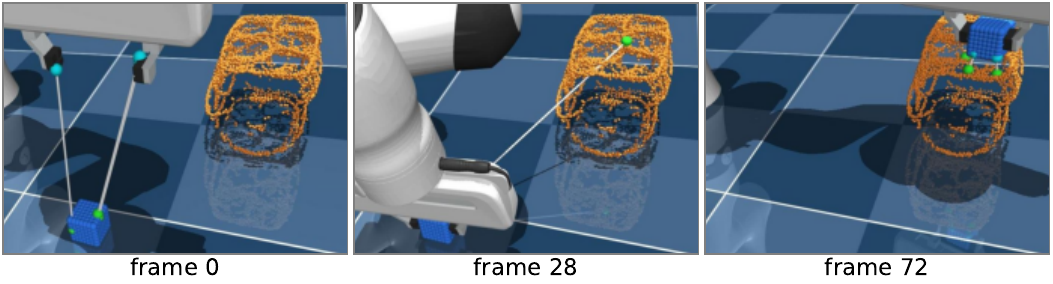}
    \caption{\textbf{Contact schedule from replaying the optimized contact
    geometry $C^*$.} Simulation frames at the three key timesteps of the rollout
    (frame index continuous across stages): \emph{frame~0} the gripper reaches
    toward the block, \emph{frame~28} the grasped block is transported toward the
    hole, and \emph{frame~72} it is aligned and inserted. The paired cyan and
    green spheres connected by white lines denote the active contact pairs of the
    contact event sequence $E$ that define the contact reward.
    }
    \label{fig:rl_schedule}
\end{figure}

\paragraph{The contact-event reward imposes the demonstrated contact schedule.}
The second component turns the optimized geometry into a learning objective.
Figure~\ref{fig:rl_schedule} visualizes the contact schedule realized when the
demonstration is replayed through the optimized contact geometry $C^*$: the
staged contact events are read off directly as the contact event sequence $E$.
This sequence $E$ is exactly what the contact-event reward enforces during
policy learning. It explains the Q1 ablation in which replacing this reward with
a hand-designed one degrades insertion: a hand-designed reward cannot reproduce
the precise, demonstration-grounded contact schedule that $E$ encodes, so the
policy drifts into unrealistic contact regimes.

Together, the two components are complementary: contact geometry optimization
makes the simulated local contact dynamics faithful, and the contact-event
reward steers the policy through the demonstrated contact schedule.

\subsection{Q3: Scalability of VCP}
\label{subsec:vcp_results}

\begin{figure}[t]
    \centering
    \includegraphics[width=0.41\textwidth]{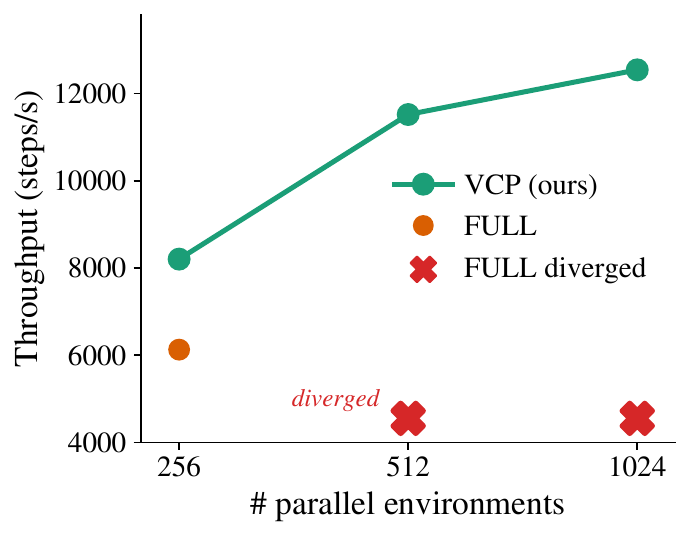}
    \caption{\textbf{Throughput scaling of VCP vs.\ FULL physical contact}.
    VCP scales steadily with the number of parallel environments, whereas FULL is stable only at 256 and diverges at $\geq\!512$ environments when the simultaneous contacts exceed the contact-buffer capacity (red $\times$).}
    \label{fig:vcp_scaling}
\end{figure}
Finally, we test whether VCP (Sec.~\ref{sec:vcp}) enables collision avoidance
without the cost of full physical contact. Keeping everything else fixed, we
change only the object contact model: FULL activates all 512 surface
spheres as MuJoCo collision geoms, whereas VCP keeps only the 78 primitives
that actually touched during the demonstration replay
(Sec.~\ref{sec:contact_extraction}) and resolves the rest with an analytic
virtual contact penalty.

VCP's throughput scales steadily with the number of parallel environments;
at 256---the only scale where FULL stays stable---VCP is already $\sim\!1.3\times$
faster (Fig.~\ref{fig:vcp_scaling}). Beyond that, FULL collapses: at
$\geq\!512$ environments the simultaneous
contacts exceed the fixed contact buffer, contacts are dropped, and training
diverges (3/3 seeds), while VCP stays stable at every scale. This fixed buffer is a structural requirement of GPU-batched simulators, which cannot use ragged arrays; VCP sidesteps it by capping contacts with a fixed-cost analytic penalty.

\section{Conclusion}
\label{sec:conclusion}

We presented a real-to-sim-to-real framework that minimizes the sim-to-real gap by constraining the RL policy's search space with the task-relevant contact dynamics
identified from a real-world demonstration, rather than by matching global physical parameters.
Concretely, we optimize the contact geometry so that the simulator physically explains the
object point cloud trajectory extracted from a single demonstration, thereby aligning the
local contact dynamics, and we replay the demonstration to automatically extract a
contact event sequence, which is used as a reward signal that guides the
policy toward physically plausible interaction regimes.
A Virtual Collision Penalty (VCP) further makes collision-avoidance learning tractable at low cost.
On a shape-sorter insertion task with a $\qty{2}{\milli\meter}$ tolerance, our method achieves an 80.0\%
cumulative insertion success rate, substantially outperforming ablations.

The broader significance of this work is the finding that task-relevant contact dynamics can
be learned from only a few demonstrations and exploited as a constraint to achieve sim-to-real policy transfer of contact-rich manipulation without large-scale real-world data.
This offers a real-world-data-efficient pathway that does not rely on large-scale imitation
learning data.
Moreover, because such contact-centric constraints are inherently manipulator-agnostic, they
suggest a promising extension toward dexterous manipulation with multi-fingered hands, which we leave for future work.

\paragraph{Limitations and Future Work}
This work has three main limitations.
First, our evaluation is confined to single rigid-body insertion with a parallel-jaw gripper;
generalization to a wider range of contact-rich tasks (e.g., multi-step assembly, deformable
objects, or multi-fingered manipulation) remains to be validated.
Second, more efficient approaches to contact geometry optimization and in-simulation object
representation merit further investigation.
Third, real-world deployment relies on a VLA distillation strategy that requires upfront rendering
and data curation, which may become a bottleneck; more efficient ways to deploy the trained RL policy directly in the real world are required.

\acknowledgments{We thank Ilia Chelak, Zeyi Huang, and Lilika Makabe for helpful discussions on Gaussian Splatting.}

\clearpage
\bibliography{reference}

\clearpage
\appendix
\section*{Appendix}
\addcontentsline{toc}{section}{Appendix}

\section{Contact Geometry Optimization: Implementation Details}
\label{sec:appendix_es}

We optimize the contact geometry $C$ using an evolutionary strategy (ES)~\citep{beyer2002evolution}---a population-based black-box optimizer that, at each iteration, samples a population of candidate perturbations, evaluates them in simulation, and moves the current estimate toward the best-scoring ones--- with spatially localized perturbations. The key design principle is that only primitives near observed contacts are adjusted, preserving the global shape while refining local collision surfaces.

\textbf{Localized deformation mask.} At each generation, we identify primitives that have participated in contacts during simulation replay. We then expand this set to its $k$-nearest neighbors, forming a \emph{deformation mask}. Only primitives within this mask are subject to perturbation; all others remain fixed.

\textbf{RBF-smoothed perturbation.} To ensure spatially coherent deformations, we construct a radial basis function (RBF) kernel over the primitive positions. Random noise is generated at the centroid of the deformable region and propagated through the RBF kernel, producing smooth, physically plausible displacements rather than independent per-primitive jitter.

\textbf{Elite selection and update.} From the population, the best-scoring candidates (elites) are selected by chamfer distance between simulated and observed point cloud trajectories. The mean position is updated via exponential moving average, again smoothed by the RBF kernel within the deformation mask.

\section{Contact Event Staging from the Demonstration}
\label{sec:appendix_staging}

The contact event sequence $E=(e_1,\dots,e_S)$ is obtained by partitioning the demonstration into $S$ contiguous stages along its contact-phase transitions. Each stage $s$ has a single target event $e_s$, the contact configuration to be achieved in that stage. During RL, every state sampled from stage $s$ is rewarded for reproducing $e_s$. This many-to-one mapping stabilizes the reward within each stage, easing RL training.

For the shape-sorter task, the contact transitions naturally separate three stages: picking begins when the gripper contacts Object~1 (the block), aligning when Object~1 contacts Object~2 (the box) around the hole, and inserting when Object~1 contacts the inner wall of the hole. Each new contact marks a stage boundary, and the contact configuration at that point becomes the stage's target event $e_s$.

\section{Contact-Centric Policy Optimization: Training Details}
\label{sec:appendix_ppo}

This section specifies the observation space, action space, and reward terms.
The observation $\mathbf{o}_t \in \mathbb{R}^{148}$ is detailed in Table~\ref{tab:obs}. It consists of a $44$-dimensional base (robot and object state) followed by $N{=}8$ fixed contact slots of $13$ dimensions each. Each slot carries a binary mask indicating whether it holds an active contact pair; unused slots are zero-padded, so the demonstration's active contacts per step occupy the first slots and the rest are padding. Each arm joint angle $\theta$ is represented as $(\sin\theta,\cos\theta)$ to avoid the $2\pi$ wrap-around discontinuity, and object orientations use the $6$D continuous rotation representation~\citep{zhou2019continuity}. Per slot, the contact features are the contact point $\mathbf{c}_t^{(i)}$, the displacement \hbox{$\boldsymbol{\delta}_t^{(i)}=\mathbf{b}_t^{(i)}-\mathbf{c}_t^{(i)}$} toward its demonstration target $\mathbf{b}_t^{(i)}$, its unit direction (the approach direction) \hbox{$\hat{\mathbf{u}}_t^{(i)}=\boldsymbol{\delta}_t^{(i)}/\lVert\boldsymbol{\delta}_t^{(i)}\rVert$}, and the target surface normal $\hat{\mathbf{n}}^{(i)}$, each $3$-dimensional.

\begin{table}[h]
\centering
\caption{Observation space $\mathbf{o}_t \in \mathbb{R}^{148}$: a $44$-dim base plus $N{=}8$ contact slots of $13$ dims each ($8\times13=104$).}
\label{tab:obs}
\begin{tabular}{l c}
\toprule
Component & Dim \\
\midrule
Arm joint angles ($\sin/\cos$ encoded) & 14 \\
Arm joint velocities & 7 \\
Gripper opening & 1 \\
Gripper command (last applied) & 1 \\
End-effector world position & 3 \\
Object~1 ($\mathrm{O}_1$) position $+$ 6D rotation & 9 \\
Object~2 ($\mathrm{O}_2$) position $+$ 6D rotation & 9 \\
\emph{Base subtotal} & 44 \\
\midrule
\multicolumn{2}{l}{\emph{Per contact slot} ($N{=}8$ slots, zero-padded):} \\
\quad active mask & 1 \\
\quad contact point $\mathbf{c}_t^{(i)}$ & 3 \\
\quad displacement $\boldsymbol{\delta}_t^{(i)}$ & 3 \\
\quad approach direction $\hat{\mathbf{u}}_t^{(i)}$ & 3 \\
\quad target normal $\hat{\mathbf{n}}^{(i)}$ & 3 \\
\emph{Slot subtotal} $\times\,8$ & $13\times8=104$ \\
\midrule
Total & $148$ \\
\bottomrule
\end{tabular}
\end{table}

The action $\mathbf{a}_t \in [-1,1]^8$ consists of 7-DoF delta-joint commands for the arm and a gripper command.

The reward is the weighted sum in \eref{eq:reward}, expanded here into its
constituent terms.
The \emph{reaching} term uses a two-scale exponential shaping
\begin{equation}
r^{\text{reach}}_t = A_{\text{far}}\exp\left(-\max_i \frac{d_t^{(i)}}{\sigma_{\text{far}}}\right)
+ A_{\text{near}}\exp\left(-\max_i{\frac{d_t^{(i)}}{\sigma_{\text{near}}}}\right),
\end{equation}
where $A_\bullet$ and $\sigma_\bullet$ are the amplitude and length-scale of each
exponential, $d_t^{(i)}$ is the distance of the $i$-th active contact pair at step
$t$, and $i$ indexes the active pairs.
A \emph{far} term (large $\sigma_{\text{far}}$) provides a dense long-range
gradient, while a \emph{near} term (small $\sigma_{\text{near}}$) sharpens precision
near contact.
The \emph{regularization} term is
\begin{equation}
r^{\text{reg}}_t = -0.01\,\lVert \mathbf{a}_t \rVert^2
- 0.001\,\lVert \dot{\mathbf{q}}_t \rVert^2 + r^{\text{disturb}}_t,
\end{equation}
where $\dot{\mathbf{q}}_t$ is the arm joint velocity. The first two terms penalize
large actions and fast joint motion to keep control smooth; $r^{\text{disturb}}_t$ penalizes
pushing or rotating the manipulated objects between steps.
The VCP term in \eref{eq:reward} is a distance-based penalty on
non-essential robot--object and object--object primitive pairs, together with gated margin-band penalties that enforce a safety
margin on near-contact pairs; a single-pair penetration beyond a threshold
terminates the episode with a fixed penalty.
The policy and value function are each parameterized as 3-layer MLPs (hidden
size 256, Swish activation), with the policy modeled as a tanh-squashed
Gaussian.

\section{Contact-Consistent Reset Strategy: Detailed Cases}
\label{sec:appendix_reset}

As described in Sec.~\ref{sec:ppo}, we diversify initial states by sampling poses from the demonstration while preserving the current contact chain. Here, Object~1 ($\mathrm{O}_1$) denotes the object grasped by the robot (in our shape-sorter task, the block), and Object~2 ($\mathrm{O}_2$) the object in contact with $\mathrm{O}_1$ (the box with the insertion hole). We consider three cases based on the contact state:
\begin{enumerate}
    \item \textbf{No robot--object contact:} The robot, $\mathrm{O}_1$, and $\mathrm{O}_2$
    are reset freely. The robot is reset in joint angle space, while $\mathrm{O}_1$ and $\mathrm{O}_2$ are reset in their $x, y$ positions and orientations.
    \item \textbf{Robot--object contact:} The relative pose between the robot end-effector and $\mathrm{O}_1$ is maintained while both are translated simultaneously. The robot joint angles are computed via inverse kinematics (IK) to satisfy the resulting end-effector pose. Resets for which IK fails are discarded.
    \item \textbf{Robot--object and object--object contact:} The robot, $\mathrm{O}_1$, and $\mathrm{O}_2$ are all translated simultaneously while maintaining their relative poses. Robot joints are solved via IK as in case~2.
\end{enumerate}

\section{Hybrid Rendering Pipeline: Architecture Details}
\label{sec:appendix_visual}

\subsection{3DGS for Objects}

We use an iPhone-based multi-view capture tool (SplatsCam) to acquire approximately 100 multi-view images per object. We then run COLMAP~\citep{schoenberger2016sfm} followed by 3DGS to train a Gaussian splat model. The primitive groups in simulation are sampled from this 3DGS model. Since 3DGS optimizes for appearance, Gaussian density tends to be disproportionately high in regions with visually complex patterns. To ensure uniform spatial coverage, we apply Farthest Point Sampling (FPS) in Euclidean space to sample primitives.

\subsection{Flow-Matching-Based Robot-Background Rendering}

We train a U-Net~\citep{ronneberger2015unet}-based flow matching~\citep{lipman2023flow} generative model for robot-background rendering, stabilized with Exponential Moving Average (EMA)~\citep{polyak1992ema}.

The U-Net input consists of a channel-wise concatenation (9 channels total) of: the noised image, a simulation-rendered image corresponding to the current joint pose, and a past real image (ground truth during training; the previously generated image in an autoregressive manner during inference). For the hand-mounted camera mode, we additionally condition on the robot end-effector pose (3D position + 6D rotation representation) via FiLM~\citep{perez2018film} conditioning added to the time embedding. The output is the denoised image.

Since our flow-matching-based rendering must generate large volumes of synthetic images for training, inference speed is critical. We therefore adopt a low-resolution flow matching combined with super-resolution approach. Images of size $256 \times 256$ are resized to $64 \times 64$ (using torchvision bilinear interpolation), and flow matching is trained and run at this resolution. A separate super-resolution network then learns the $64 \times 64 \rightarrow 256 \times 256$ mapping. Following EDSR~\citep{lim2017edsr}, the architecture comprises residual blocks without BatchNorm~\citep{ioffe2015batchnorm}, a $4\times$ PixelShuffle~\citep{shi2016pixelshuffle} upsampler, and a global residual skip via bicubic upsampling, trained with L1 loss in a single forward pass. This approach yields approximately $10\times$ speedup compared to operating at $256 \times 256$ directly.

\subsection{Shadow Augmentation}

Flow matching can model appearance factors between the robot and background, such as the robot's shadow cast on the table. However, since objects are rendered separately via Gaussian splatting, appearance interactions with objects (e.g., object shadows) cannot be captured. To mitigate this, we approximately model shadows cast by objects onto the table as a rendering augmentation. Specifically, when object primitive groups are posed in the training dataset, we randomly sample virtual light sources and compute the shadow projection of each object onto the ground plane ($z = 0$\,mm).

\subsection{Qualitative Rendering Samples}
\label{sec:appendix_render_samples}

\begin{figure}[t]
    \centering
    \includegraphics[width=\textwidth]{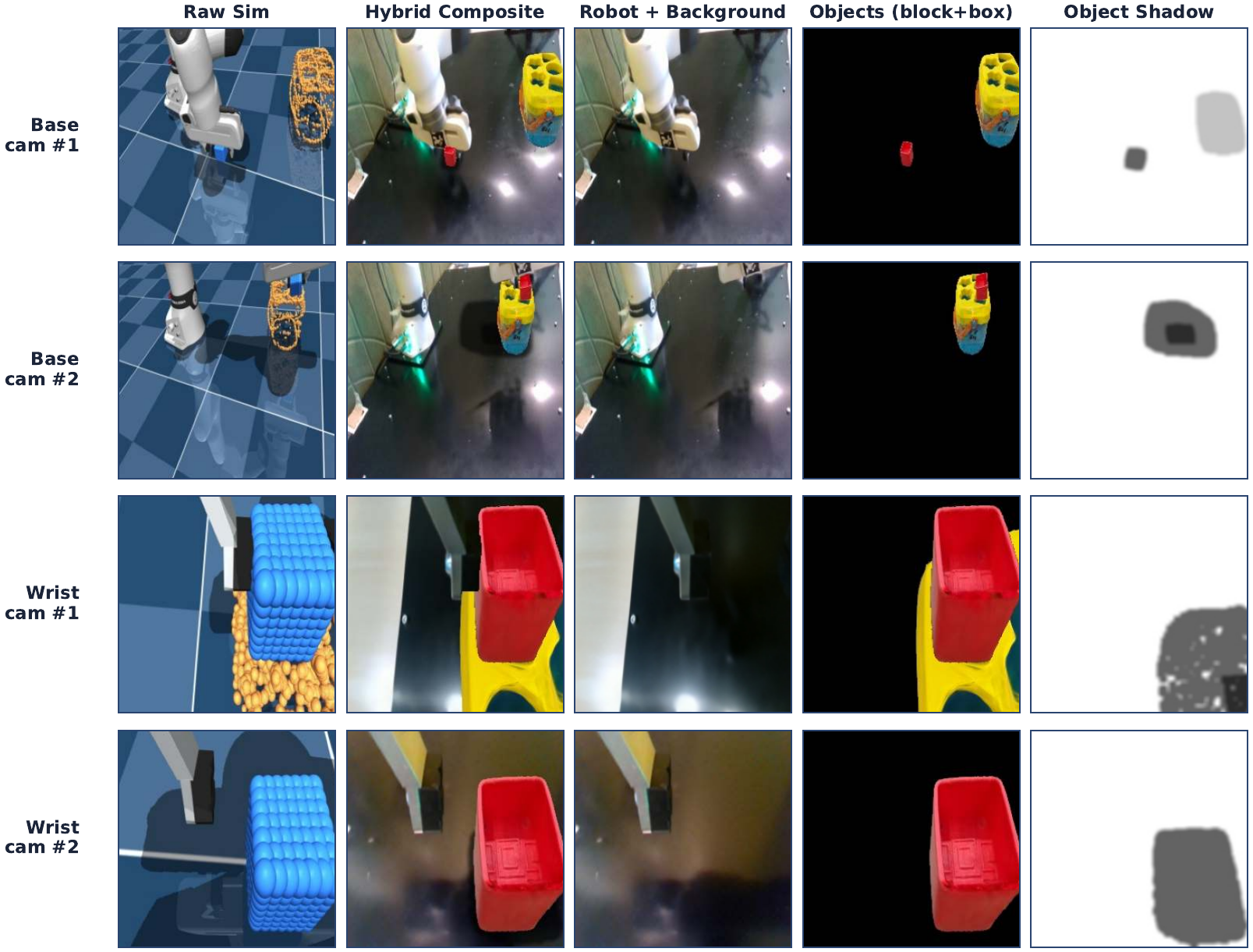}
    \caption{\textbf{Decomposition of the hybrid rendering pipeline.}
    Rows: two base-camera and two wrist-camera samples;
    columns: raw sim render $\mid$ final hybrid composite $\mid$
    flow-matching robot-and-background $\mid$ 3DGS objects (block$+$box) $\mid$
    object-shadow render. The last three columns are composited into the second,
    closing the sim-to-real appearance gap used for VLA distillation.}
    \label{fig:hybrid_samples}
\end{figure}

Figure~\ref{fig:hybrid_samples} decomposes the hybrid rendering pipeline. For
each pose we show the raw MuJoCo render, the final hybrid composite, and the
three layers that compose it: the flow-matching robot-and-background render, the
3DGS object render, and the
augmented object-shadow render. Compositing these three layers yields the
photorealistic image used for VLA distillation, while the raw render exposes the
underlying contact primitives.

\section{Reward Ablation: Engineered Insertion Reward Baseline}
\label{sec:appendix_engineered_reward}

To isolate the effect of the proposed contact-based reward (contact-event reward), we
construct a baseline that replaces only the insertion-stage reward with a hand-designed
(engineered), geometry-based reward.
This baseline inherits the proposed method's training pipeline unchanged, and replaces only
the reward terms as follows:
\begin{equation}
    r_t^{\text{eng}} = r_t^{\text{align-xy}} + r_t^{\text{depth-z}} + r_t^{\text{success}}
    + r_t^{\text{reg}} + r_t^{\text{vcp}}.
    \label{eq:engineered_reward}
\end{equation}

The hole center is given as a manually specified offset
$\mathbf{p}^{\text{hole}}_{\text{local}}$ in the local frame of the target object
$\mathrm{O}_2$
, and at each step moves with the actual pose of $\mathrm{O}_2$ as $\mathbf{p}^{\text{hole}}_{t} = \mathbf{p}^{\mathrm{O}_2}_t + R^{\mathrm{O}_2}_t\, \mathbf{p}^{\text{hole}}_{\text{local}}$.
Each reward term is defined as follows.
\begin{itemize}
    \item \textbf{XY alignment} $r^{\text{align-xy}}$: a \emph{dual-scale} exponential
    reward on the horizontal distance $d_{xy}$ between the insertion block
    $\mathrm{O}_1$ and the hole center,
    $A_f \exp(-d_{xy}/\sigma_f) + A_n \exp(-d_{xy}/\sigma_n)$
    (far term $\sigma_f{=}\qty{100}{\milli\meter}$, near term $\sigma_n{=}\qty{10}{\milli\meter}$).
    \item \textbf{Z insertion depth} $r^{\text{depth-z}}$: a dual-scale exponential
    reward on the height remaining above the hole entrance
    $z^{+} = \max(z_{\mathrm{O}_1} - z^{\text{target}},\,0)$, where $z^{\text{target}}$ is the world height the block must reach to be fully seated in the hole. It is multiplied by a
    sigmoid gate that turns on near $d_{xy}{=}\qty{30}{\milli\meter}$ so that it is active only
    when $XY$ is aligned.
    \item \textbf{Success bonus} $r^{\text{success}}$: a large constant bonus ($50$) is
    granted when $d_{xy}$ is within a threshold ($\qty{10}{\milli\meter}$) and
    $z_{\mathrm{O}_1}$ has descended to $z^{\text{target}}$ or below.
\end{itemize}
This baseline requires \emph{per-task manual design} of (i) the hole position $\mathbf{p}^{\text{hole}}_{\text{local}}$ and the target seating height $z^{\text{target}}$, (ii) the amplitude/scale of each term,
and (iii) the success threshold; this very reward engineering is exactly the cost that
the proposed contact-based reward aims to remove.

\end{document}